\documentclass[twoside]{article}

\usepackage[accepted]{aistats2022}
\usepackage{xcolor}
\usepackage{bbold}
\usepackage{enumitem}
\usepackage{mystyle}
\usepackage{adjustbox}
\usepackage{hyperref}
\usepackage{xargs} 
\usepackage{makecell}
\usepackage{amssymb,amsmath}
\usepackage{amsthm}
\usepackage{mathtools}
\mathtoolsset{showonlyrefs}

\usepackage[round]{natbib}

\newcommand{\e}{\varepsilon}
\newcommand{\sm}{\texttt{SoftMax}}
\newcommand{\sk}{\texttt{Sinkhorn}}
\begin{document}

\twocolumn[

\aistatstitle{Sinkformers: Transformers with Doubly Stochastic Attention}

\aistatsauthor{ Michael E. Sander \And Pierre Ablin  \And Mathieu Blondel \And  Gabriel Peyré}

\aistatsaddress{ENS and CNRS \And  ENS and CNRS \And Google Research, Brain team \And ENS and CNRS}]

\begin{abstract}
Attention based models such as Transformers involve pairwise interactions between data points, modeled with a learnable attention matrix. Importantly, this attention matrix is normalized with the SoftMax operator, which makes it row-wise stochastic. 
In this paper, we propose instead to use Sinkhorn's algorithm to make attention matrices doubly stochastic. We call the resulting model a Sinkformer. We show that the row-wise stochastic attention matrices in classical Transformers get close to doubly stochastic matrices as the number of epochs increases, justifying the use of Sinkhorn normalization as an informative prior.
On the theoretical side, we show that, unlike the SoftMax operation, this normalization makes it possible to understand the iterations of self-attention modules as a discretized gradient-flow for the Wasserstein metric. 
We also show in the infinite number of samples limit that,  when rescaling both attention matrices and depth, Sinkformers operate a heat diffusion. 
On the experimental side, we show that Sinkformers enhance model accuracy in vision and natural language processing tasks. 
In particular, on 3D shapes classification, Sinkformers lead to a significant improvement.
\end{abstract}

\section{Introduction} %

The Transformer \citep{vaswani2017attention}, an architecture that relies entirely on attention mechanisms \citep{bahdanau_2014}, has achieved state of the art empirical success in natural language processing (NLP) \citep{brown2020language, radford2019language, wolf2019huggingface} as well as in computer vision \citep{dosovitskiy2020image, zhao2020point, zhai2021scaling, lee2019set}. As the key building block of the Transformer, the self-attention mechanism takes the following residual form \citep{yun2019transformers} given
a $n$-sequence $(x_1 , x_2 , ... , x_n)$, embedded in dimension $d$:
\begin{equation}\label{eq:attention}
    x_i \leftarrow x_i +\sum^{n}_{j=1}  K^{1}_{i,j}  W_V x_j,
\end{equation}
where ${K^{1}} \coloneqq \sm(C)$ with $C_{i,j} \coloneqq ({W_Qx_i})^{\top}{W_Kx_j}$
$= x_i^\top W_Q^\top W_K x_j$.
Here, $W_Q, W_K \in \RR^{m\times d}$ and $ W_V \in \RR^{d\times d}$ are the query, key and value matrices.
The SoftMax operator can be seen as a normalization of the matrix $K^0 \coloneqq \exp(C)$ as follows: $K^1_{ij} \coloneqq K^0_{ij}/ \sum_{l=1}^n K^0_{il}$ for all $i$ and $j$.
Importantly, the matrix $K^{1}$  is row-wise stochastic: its rows all sum to $1$. 

\begin{figure}[H]
\centering
\includegraphics[width=\columnwidth]{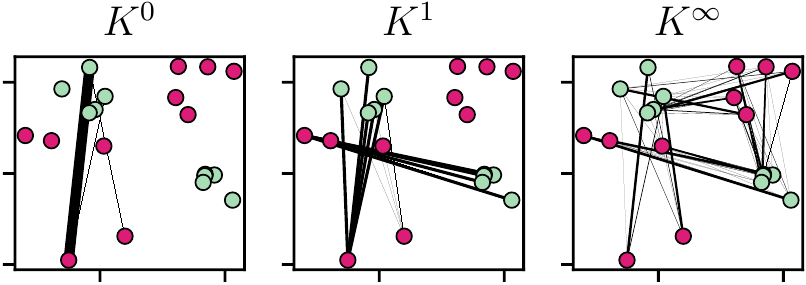} 
\caption{\textbf{Illustration of the different normalizations of attention matrices.}
We form two point clouds $(W_Qx_i)_{1\leq i \leq 10}$ (green) and $(W_Kx_j)_{1\leq i \leq 10}$ (red). For $k \in \{0, 1, \infty\}$, the width of the line connecting $x_i$ to $x_j$ is $K^{k}_{i,j}$. We only display connections with $K^{k}_{i,j} \geq 10^{-12}$. For $K^{0}$, one interaction dominates. For $K^{1}$ (SoftMax), one cluster is ignored. For $K^{\infty}$ (Sinkhorn), all points are involved in an interaction.
}\label{fig:different_norm}
\vspace{-1em}
\end{figure}

In this work, we propose to take the normalization  process further by successively normalizing the rows and columns of $K^{0}$.
This process is known to provably converge to a doubly stochastic matrix 
(i.e., whose rows and columns both sum to $1$)
and is called Sinkhorn's algorithm \citep{sinkhorn1964relationship, cuturi2013sinkhorn, peyre2019computational}. We denote the resulting doubly stochastic matrix $K^{\infty}$. Intuitively, such a normalization relies on a democratic principle where all points are matched one to another with different degrees of intensity, so that more interactions are considered than with the SoftMax normalization, as shown in Figure~\ref{fig:different_norm}.
\vspace{-1em}

We call our Transformer variant where the SoftMax is replaced by Sinkhorn a \textbf{Sinkformer}.
Since Sinkhorn's first iteration coincides exactly with the SoftMax,
Sinkformers include Transformers as a special case. 
Our modification is differentiable, easy to implement using deep learning libraries, and can be executed on GPUs for fast computation. 
Because the set of row-wise stochastic matrices contains the set of doubly stochastic matrices, the use of doubly stochastic matrices can be interpreted as a prior.
On the experimental side, we confirm that doubly stochastic attention leads to better accuracy in several learning tasks. On the theoretical side, doubly stochastic matrices also give a better understanding of the mathematical properties of self-attention maps.

To summarize, we make the
following contributions.
\begin{itemize}[topsep=0pt,itemsep=2pt,parsep=2pt,leftmargin=10pt]

\item We show empirically that row-wise stochastic matrices seem to converge to doubly stochastic matrices during the learning process in several classical Transformers (Figure \ref{fig:sum_training}). Motivated by this finding, we then introduce the Sinkformer, an extension of the Transformer in which the SoftMax is replaced by the output of Sinkhorn's algorithm. In practice, our model is parametrized by the number of iterations in the algorithm, therefore interpolating between the Transformer and the Sinkformer.

\item On the theoretical side, we show that Transformers and Sinkformers can be viewed as models acting on discrete distributions, and we show under a symmetry assumption that Sinkformers can be seen in the infinite depth limit as a Wasserstein gradient flow for an energy minimization (Proposition \ref{prop:gradient_flows}). We also show that the classical Transformer with the SoftMax operator cannot be interpreted as such a flow (Proposition \ref{prop:not_a_gradient}). To the best of our knowledge, this is the first time such a connection is established.
We also prove that in the infinite number of particles limit (when $n$ goes to infinity), the iterations of Sinkformers converge to the heat equation (Theorem \ref{thm:diffusion}), while the corresponding equation for Transformers is nonlinear and nonlocal (Proposition \ref{prop:soft}).

\item On the experimental side, we show that Sinkformers lead to a significant accuracy gain compared to Transformers on the ModelNet 40 3D shapes classification task.
We then demonstrate better performance of Sinkformers on the NLP IMDb dataset for sentiment analysis and IWSLT'14 German to English neural machine translation tasks. Sinkformers also achieve a better accuracy than Vision Transformers on image classification tasks. Therefore, the proposed method is capable of enhancing the performance of transformers in a wide range of applications.

\end{itemize}

\section{Background and related work}\label{sec:background}

\paragraph{Transformers.} Proposed by \cite{vaswani2017attention}, the Transformer is a fully attention-based architecture. Originally designed to process sequences for natural language processing (NLP), many variants have since been developed such as Vision Transformers \citep{dosovitskiy2020image, zhai2021scaling}, Set Transformers \citep{lee2019set} or Point Cloud Transformers \citep{zhao2020point}. The Transformer and its variants are based on an encoder-decoder structure, where the decoder can have a more or less complex form. The encoder is fully \textit{self}-attention based. After embedding and concatenating with positional encoding the original input sequence, the encoder uses a series of residual blocks that iterates relation \eqref{eq:attention} followed by a feed forward neural network applied to each $x_i$ independently.
In its most complex form such as in neural machine translation, the decoder combines a \text{self}-attention based mechanisms and a \textit{cross} attention one, meaning that it is given access to the encoder via another multi-head attention block. 

\paragraph{Sinkhorn and Attention.} 

To the best of our knowledge, using Sinkhorn's algorithm in Transformers has been done once in a different context \citep{tay2020sparse}. 
The authors  propose to learn efficient and sparse attention using a differentiable algorithm for sorting and rearranging elements in the input sequence. For this purpose, they introduce a sorting network to generate a doubly-stochastic matrix (that can be seen as a relaxed version of a permutation matrix) and use it to sort the sequence in a differentiable fashion. \citet{mialon2021trainable} propose an embedding for sets of features in $\RR^d$ based on Sinkhorn's algorithm, by using the regularized optimal transport plan between data points and a reference set. 
\citet{niculae_2018} use doubly stochastic attention matrices in LSTM-based encoder-decoder networks but they use Frank-Wolfe or active set methods to compute the attention matrix.
None of these works use Sinkhorn on self-attention maps in Transformers and provide its theoretical analysis, as we do. 

\paragraph{Impact of bi-normalization.}
Theoretical properties of kernels $\Kk$, which attention is an instance of, can also be studied through the operator $f \mapsto f - \Kk f$. 
Bi-normalization of kernels over manifolds have already been studied in the literature, on uniform measures  \citep{singer2006graph}, weighted measures \citep{hein2007graph} and in a more general setup with associated diffusion operators \citep{ting2011analysis}. \citet{milanfar2013symmetrizing} proposes to approximate smoothing operators by doubly stochastic matrices using Sinkhorn's updates, leading to better performance in data analysis and signal processing. Importantly, the works of \cite{marshall2019manifold} and \cite{wormell2021spectral} exactly introduce a normalization that is based on Sinkhorn's algorithm. They prove that this method models a Langevin diffusion and leads to the approximation of a symmetric operator. They also show that convergence to this operator is faster with Sinkhorn normalization than with the SoftMax normalization. In section \ref{sec:laplacians}, we adopt a similar point of view with a parametrized cost and show that different normalizations result in different partial differential equations (PDEs) in the infinite number of particles limit.

\paragraph{Infinite depth limit.}

 Studying deep residual neural networks (ResNets) \citep{he2016deep} in the infinitesimal step-size regime (or infinite depth limit) has recently emerged as a new framework for analyzing their theoretical properties. The ResNet equation
 \begin{equation}\label{eq:resnet}
     x_i \leftarrow x_i + T(x_i)
 \end{equation}
can indeed be seen as a discretized Euler scheme with unit step size of the ordinary differential equation (ODE) $\dot{x_i}=T(x_i)$ \citep{E_2017,chen2018neural,dupont2019augmented,sun2018stochastic,E_2018,lu2017finite,ruthotto2018deep, pmlr-v139-sander21a}. In section \ref{sec:gradient_flows}, we adopt this point of view on residual attention layers in order to get a better theoretical understanding of attention mechanisms. This is justified by the fact that, for instance, GPT-3 \citep{brown2020language} has 96 layers. 
\paragraph{Neural networks on measures.}
The self-attention mechanism \eqref{eq:attention} acts on sets $\{ x_i \}_{i}$ where the ordering of the elements does not matter. An equivalent way to model such invariant architectures is to consider them as acting on probability measures or point clouds of varying cardinality \citep{de2019stochastic, vuckovic2021regularity, zweig2021functional}.
Specifically, a collection of points $(x_i)_{1\leq i \leq n}$, where $x_i \in \RR^{d}$, can also be seen as a discrete measure on $\RR^d$: $\mu \coloneqq \frac{1}{n} \sum_{i=1}^{n} \delta_{x_i} \in \Mm(\RR^d)$,
where $\Mm(\RR^d)$ is the set of probability measures on $\RR^d$. 
A map $T_\mu$ then acts on $\mu$ through $F(\mu) \coloneqq \frac{1}{n} \sum_{i=1}^{n} \delta_{T_\mu(x_i)}$. One notable interest of such a point of view is to consider the evolution of non ordered sets of points. Another is to consider the mean field (or large sample) limit, that is when $n \to \infty$, to conduct theoretical analysis \citep{zweig2021functional} as when analyzing the SGD properties in the mean-field limit \citep{song2018mean}.

\section{Sinkformers}

We now introduce Sinkformers, a modification of any Transformer by replacing the SoftMax operator in the attention modules by Sinkhorn's algorithm. 

\begin{figure}[h]
\centering
\includegraphics[width=\columnwidth]{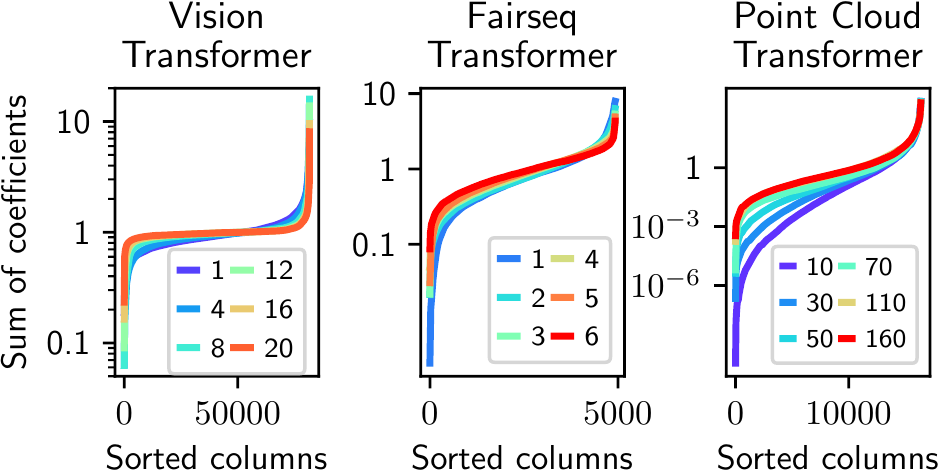} 
\caption{\textbf{Sum over columns} of attention matrices at different training epochs (color) when training, from left to right, a ViT on MNIST (section \ref{sec:vit}), a \texttt{fairseq} Transformer on IWSLT'14 (section \ref{sec:fair}), and a Point Cloud Transformer on Model Net 40 (section \ref{sec:modelnet}).
\textbf{The majority of columns naturally sum closely to $1$.}
}\label{fig:sum_training}
\vspace{-1em}
\end{figure}

\paragraph{Attention matrices during training.}
In Transformers, attention matrices are row-wise stochastic. A natural question is how the sum over columns evolve during training. On $3$ different models and $3$ different learning tasks, we calculated the sum over columns of attention matrices in Transformers. We find out that the learning process makes the attention matrices more and more doubly stochastic, as shown in Figure \ref{fig:sum_training}. 

Thus, row-wise stochastic attention matrices seem  to  approach  doubly  stochastic  matrices  during  the  learning  process  in classical  Transformers. Therefore, it seems natural to impose double stochasticity as a prior and study theoretically and experimentally the resulting model. A process to obtain such matrices which extends the SoftMax is Sinkhorn's algorithm.

\paragraph{Sinkhorn's algorithm.}

Given a matrix $C \in \RR^{n \times n}$, and denoting $K^{0} \in \RR^{n \times n}$ such that $K^{0} = \exp(C)$, Sinkhorn's algorithm \citep{sinkhorn1964relationship, cuturi2013sinkhorn, peyre2019computational} iterates, starting from $K^{0}$:
\begin{equation}\label{eq:sinkhorn_alg}
K^{l+1} = 
\left\{
\begin{array}{r@{\hspace{1mm}}l}
&N_R(K^{l})\quad\text{if $l$ is even} \\
&N_C(K^{l})\quad\text{if $l$ is odd},
\end{array}
\right.
\end{equation}
where $N_R$ and $N_C$ correspond to row-wise and column-wise normalizations: $(N_R(K))_{i,j} \coloneqq \frac{K_{i,j}}{\sum_{l=1}^{n} K_{i,l}}$ and $(N_C(K))_{i,j} \coloneqq \frac{K_{i,j}}{\sum_{l=1}^{n} K_{l,j}}$.
We denote the resulting scaled matrix limit $K^{\infty} \coloneqq \sk(C)$. Note that it is doubly stochastic in the sense that $ K^{\infty} \mathbb{1}_n = \mathbb{1}_n$ and ${K^{\infty}}^{\top} \mathbb{1} _{n} = \mathbb{1}_n$. The operations in \eqref{eq:sinkhorn_alg} are perfectly suited for being executed on GPUs \citep{charlier2021kernel, cuturi2013sinkhorn}.

\paragraph{Sinkformers.}

For simplicity, we consider a one head attention block that iterates equation \eqref{eq:attention}.
Note that $K^1 \coloneqq \sm(C)$ is  precisely the output of Sinkhorn's algorithm \eqref{eq:sinkhorn_alg} after $1$ iteration. 
In this paper, we propose to take Sinkhorn's algorithm several steps further until it approximately converges to a doubly stochastic matrix $K^{\infty}$. This process can be easily implemented in practice, simply by plugging Sinkhorn's algorithm into self-attention modules in existing architectures, without changing the overall structure of the network. We call the resulting drop-in replacement of a Transformer a Sinkformer. 
It iterates
\vspace{-1em}
\begin{equation}\label{eq:attention_particles_sinkhorn}
    x_i \leftarrow x_i + \sum^{n}_{j=1}  K^{\infty}_{i,j}  W_V x_j.
\end{equation}
In the next two sections \ref{sec:gradient_flows} and \ref{sec:laplacians}, we investigate the theoretical properties of Sinkformers. We exhibit connections with energy minimization in the space of measures and the heat equation, thereby proposing a new framework for understanding attention mechanisms.
All our experiments are described in Section \ref{sec:experiments} and show the benefits of using Sinkformers in a wide variety of applications. 

\paragraph{Computational cost and differentiation.} 

Turning a Transformer into a Sinkformer simply relies on replacing the SoftMax by Sinkhorn, i.e., substituting $K^1$ with $K^\infty$. In practice, we use a finite number of Sinkhorn iterations and therefore use $K^{l}$, where $l$ is large enough so that $K^{l}$ is almost doubly stochastic.
Doing $l$ iterations of Sinkhorn takes $l$ times longer than the SoftMax. However, this is not a problem in practice because Sinkhorn is not the main computational bottleneck and because only a few iterations of Sinkhorn are sufficient (typically $3$ to $5$) to converge to a doubly stochastic matrix. As a result, the practical training time of Sinkformers is comparable to regular Transformers, as detailed in our experiments.

Sinkhorn is perfectly suited for backpropagation (automatic differentiation), by differentiating through the operations of \eqref{eq:sinkhorn_alg}. The Jacobian of an optimization problem solution can also be computed using the implicit function theorem \citep{griewank_2008, krantz_2012, blondel2021efficient} instead of backpropagation if the number of iterations becomes a memory bottleneck. Together with Sinkhorn, implicit differentiation has been used by \cite{luise_2018} and \cite{cuturi_2020}.
 
\paragraph{Invariance to the cost function.}

Recall that in practice one has $C_{i,j} = ({W_Qx_i})^{\top}{W_Kx_j}$. An important aspect of Sinkformers is that their output is unchanged if the cost is modified with non interacting terms, as the next proposition shows. 
\begin{prop}\label{prop:modularity}
    Let $C \in \RR^{n\times n}$. Consider, for $(f, g) \in \RR^n \times \RR^n$ the modified cost function $\Tilde{C}_{i,j} \coloneqq C_{i,j} + f_i + g_j$. Then $\sk(C) = \sk(\Tilde{C})$.
\end{prop}
A proof is available in Appendix \ref{app:proofs}. A consequence of this result is that one can consider the cost $\Tilde{C}_{i,j} \coloneqq - \frac{1}{2}\| {W_Qx_i} - {W_Kx_j}\|^2$
instead of $C_{i,j} = ({W_Qx_i})^{\top}{W_Kx_j}$,
without affecting $K^\infty$. A Transformer using the cost $\Tilde{C}$ is referred to as L2 self-attention, and is Lipschitz under some assumptions \citep{kim2021lipschitz} and can therefore be used as an invertible model \citep{behrmann2019invertible}. For instance, we use $\tilde{C}$ in Proposition \ref{prop:soft}.

\section{Attention and gradient flows}\label{sec:gradient_flows}

In this section, we make a parallel between self-attention modules in Sinkformers and gradient flows in the space of measures.
We denote $\Mm(\RR^d)$ the probability measures on $\RR^d$ and $\Cc(\RR^d)$ the continuous functions on $\RR^d$.
We denote $\nabla$ the gradient operator, $\mathrm{div}$ the divergence, and $\Delta$ the Laplacian, that is $\Delta = \mathrm{div}(\nabla)$.

\paragraph{Residual maps for attention.}
We consider a one-head attention block operating with different normalizations. 
We consider the continuous counterparts of the attention matrices seen in the previous section. We denote $c(x, x') \coloneqq {(W_Qx)}^{\top}W_Kx'$ and $k^{0} \coloneqq \exp(c)$.  For some measure $\mu \in \Mm(\RR^d)$, we define the $\sm$ operator on the cost $c$ by $k^{1}(x, x') = \sm(c)(x, x') \coloneqq \frac{k^{0}(x, x')}{\int k^{0}(x, y) d \mu(y)} $. 
Similarly, we define Sinkhorn's algorithm as the following iterations, starting from $k^{0} = \exp(c)$:
\begin{equation}\label{eq:sinkhorn_alg_measure}
k^{l+1}(x,x') = 
\left\{
\begin{array}{r@{\hspace{2mm}}l}
&\frac{k^{l}(x, x')}{\int k^{l}(x, y) d \mu(y)}\quad\text{if $l$ is even} \\
&\frac{k^{l}(x, x')}{\int k^{l}(y, x) d \mu(y)}\quad\text{if $l$ is odd}.
\end{array}
\right.
\end{equation}
We denote $k^{\infty} \coloneqq \sk(c)$ the resulting limit. Note that if $\mu$ is a discrete measure supported on a $n$ sequence of particles $(x_1 , x_2 , ... , x_n)$, $\mu =\frac{1}{n} \sum_{i=1}^{n}\delta_{x_i}$, then for all $(i,j)$, $k^{0}(x_i, x_j) = K^0_{i,j}$, $k^{1}(x_i, x_j) = K^{1}_{i,j}$ and $k^{\infty}(x_i, x_j) = K^{\infty}_{i,j}$, so that $k^0$, $k^1$ and $k^\infty$ are indeed the continuous equivalent of the matrices $K^0$, $K^1$ and $K^\infty$ respectively.
\paragraph{Infinitesimal step-size regime.} In order to better understand the theoretical properties of attention matrices in Transformers and Sinkformers, we omit the feed forward neural networks acting after each attention block. We consider a succession of attention blocks with tied weights between layers and study the infinite depth limit where the output is given by solving a neural ODE \citep{chen2018neural}. 
In this framework, iterating the Transformer equation \eqref{eq:attention}, the ResNet equation \eqref{eq:resnet} and the Sinkformer equation \eqref{eq:attention_particles_sinkhorn} corresponds to a Euler discretization with step-size $1$ of the ODEs
\vspace{-1em}
\begin{equation}
\dot x_i = T_{\mu}(x_i) ~ \text{for all } i,
\end{equation} 
where $x_i(t)$ is the position of $x_i$ at time $t$.
For an arbitrary measure $\mu \in \Mm(\RR^d)$, these ODEs can be equivalently written as a continuity equation \citep{renardy_2006}
\begin{equation}\label{eq:partial}
    \partial_t\mu + \mathrm{div}(\mu T_{\mu}) = 0.
\end{equation}
When $T_\mu$ is defined by the ResNet equation \eqref{eq:resnet}, $T_{\mu} = T$ does not depend on $\mu$. It defines an advection equation where the particles do not interact and evolve independently.
When $T_{\mu}$ is defined by the Transformer equation~\eqref{eq:attention} or Sinkformer equation~\eqref{eq:attention_particles_sinkhorn}, $T_{\mu}$ has a dependency in $\mu$ and the particles interact: the local vector field depends on the position of the other particles.
More precisely we have in this case $T^1_{\mu}(x) = \int k^1(x,x')W_V x' d \mu(x')$ for the Transformer and $ T^\infty_{\mu}(x) = \int k^{\infty}(x,x')W_V x' d \mu(x')$ for the Sinkformer. It is easily seen that when $\mu$ is discrete we recover the operators in equation~\eqref{eq:attention} and~\eqref{eq:attention_particles_sinkhorn}.

\paragraph{Wasserstein gradient flows.} A particular case of equation \eqref{eq:partial} is when $T_{\mu}$ is a gradient with respect to the Wasserstein metric $W_2$.
Let $\Ff$ be a function on $\Mm(\RR^d)$. As is standard, we suppose that $\Ff$ admits a first variation at all $\mu$: there exists a function $\frac{\delta \Ff}{\delta \mu}(\mu)$ such that $\frac{d}{d \varepsilon} \Ff(\mu + \varepsilon \rho)_{|\varepsilon = 0} = \int \frac{\delta \Ff}{\delta \mu}(\mu) d \rho $ for every perturbation $\rho$ \citep{santambrogio2017euclidean}. The Wasserstein gradient of $\Ff$ at $\mu$ is then $\nabla_W \Ff (\mu) \coloneqq \nabla (\frac{\delta \Ff}{\delta \mu}(\mu))$.
The minimization of $\Ff$ on the space of measures corresponds to the PDE~\eqref{eq:partial} with $T_\mu=- \nabla_W \Ff (\mu)$.
This PDE can be interpreted as ruling the evolution of the measure $\mu$ of particles initially
distributed according to some measure $\mu_0$, for which the positions $x(t)$ follow the flow $\dot{x} = - \nabla_W \Ff (\mu)(x)$, that minimizes the global energy $\Ff$. It corresponds to a steepest descent in Wasserstein space \citep{jordan1998variational}. In Proposition \ref{prop:gradient_flows}, we show in the symmetric kernel case that Sinkformers correspond to a Wasserstein gradient flow for some functional $\Ff^\infty$, while Transformers do not.

\paragraph{Particular case.} 

An example is when $T_\mu$ does not depend on $\mu$ and writes $T_\mu = - \nabla E$ where $E : \RR^d \to \RR$. Under regularity assumptions, a solution of \eqref{eq:partial} then converges to a local minimum of $E$. This fits in the implicit deep learning framework \citep{bai2019deep}, where a neural network is seen as solving an optimization problem. 
A typical benefit of implicit models is that the iterates $x_i$ do not need to be stored during the forward pass of the network because gradients can be calculated using the implicit function theorem: it bypasses the memory storage issue of GPUs \citep{wang2018superneurons,peng2017large,zhu2017unpaired} during automatic differentiation. Another application is to consider neural architectures that include an argmin layer, for which the output is also formulated as the solution of a nested optimization problem \citep{agrawal2019differentiable, gould2016differentiating, gould2019deep}.

\paragraph{Flows for attention.}
Our goal is to determine the PDEs \eqref{eq:partial} defined by the proposed attention maps. We consider the symmetric case, summarized by the following assumption:
\begin{asp}\label{asp:sym}
${W_K}^{\top} W_Q = {W_Q}^{\top} W_K =- W_V$
\end{asp}
Assumption \ref{asp:sym} means we consider symmetric kernels (by imposing ${W_K}^{\top} W_Q = {W_Q}^{\top} W_K$), and that when differentiating $x \mapsto \exp(c(x, x'))$, we obtain $-\exp(c)W_V$.
We show that, under this assumption, the PDEs defined by $k^{0}$ and $k^{\infty}$ correspond to Wasserstein gradient flows, whereas it is not the case for $k^{1}$.
A particular case of imposing  ${W_K}^{\top} W_Q = {W_Q}^{\top} W_K$ is when $W_Q = W_K$. This equality setting is studied by \cite{kim2021lipschitz}, where the authors show that it leads to similar performance for Transformers. Since imposing ${W_K}^{\top} W_Q = {W_Q}^{\top} W_K$ is less restrictive, it seems to be a natural assumption. Imposing $W_Q^\top W_K = -W_V$ is more restrictive, and we detail the expressions for the PDEs associated to $k^0, k^1, k^{\infty}$ without this assumption in Appendix \ref{app:proofs}.
We have the following result. 
\begin{prop}[PDEs associated to $k^0, k^1, k^{\infty}$]\label{prop:gradient_flows}
Suppose Assumption \ref{asp:sym}. 
Let $\Ff^0$ and $\Ff^\infty$ $ : \Mm(\RR^d) \to \RR$ be such that $\Ff^0(\mu) \coloneqq \frac{1}{2} \int k^{0} d (\mu\otimes \mu) $ and $\Ff^\infty(\mu) \coloneqq -\frac{1}{2}\int k^{\infty} \log(\frac{k^{\infty}}{k^{0}})d (\mu\otimes \mu)$.
Then $k^0$, $k^1$ and $k^{\infty}$ respectively generate the PDEs $\frac{\partial \mu}{\partial t} + \mathrm{div}(\mu T_{\mu}^{k}) = 0$ with
$T_{\mu}^{0} \coloneqq - \nabla_W \Ff^0 (\mu)$, 
$T_{\mu}^{1} \coloneqq - \nabla[\log(\int k^{0}(\cdot, x') d \mu(x'))]$ and
$T_{\mu}^{\infty} \coloneqq - \nabla_W \Ff^\infty (\mu)$.
\end{prop}
A proof is given in Appendix \ref{app:proofs}.
Proposition \ref{prop:gradient_flows} shows that $k^0$ and $k^{\infty}$ correspond to Wasserstein gradient flows. In addition, the PDE defined by $k^1$ does not correspond to such a flow. More precisely, we have the following result. 
\begin{prop}[The SoftMax normalization does not correspond to a gradient flow]\label{prop:not_a_gradient}
One has that $T_{\mu}^{1} = -\nabla[\log(\int k^{0}(\cdot, x') d \mu(x'))]$ is not a Wasserstein gradient. 
\end{prop}
A proof is given in Appendix \ref{app:proofs}, based on the lack of symmetry of $T_\mu^1$.
As a consequence of these results, we believe this variational formulation of attention mechanisms for Sinkformers (Proposition \ref{prop:gradient_flows}) provides a perspective for analyzing the theoretical properties of attention-based mechanisms in light of Wasserstein gradient flow theory \citep{santambrogio2017euclidean}. Moreover, it makes it possible to interpret Sinkformers as argmin layers, which is promising in terms of theoretical and experimental investigations, and which is not possible for Transformers, according to Proposition \ref{prop:not_a_gradient}. %

\textcolor{black}{Our results are complementary to the one of \cite{dong2021attention}, where the authors show that, \textbf{with no skip connections} and without the feed forward neural network acting after each attention block, the output of a Transformer converges doubly exponentially with depth to a rank-1 matrix. 
On the contrary, we propose a complementary analysis by taking skip-connections into account, as is standard in Transformers. Precisely because we consider such connections, we end up with very different behaviors.
Indeed, as shown in the next section, our analysis reveals that the relative signs for $W_K$, $W_Q$ and $W_V$ imply very different behavior, such as aggregation or diffusion. The dynamics obtained when considering skip connections are therefore richer than a rank collapse phenomenon.}
\section{Attention and diffusion}\label{sec:laplacians}

In this section, we use the same notations as in section \ref{sec:gradient_flows}. We consider the mean-field limit, where the measure $\mu$ has a density with respect to the Lebesgue measure. We are interested in how the density of particles evolves for an infinite depth self-attention network with tied weights between layers. We consider Assumption \ref{asp:sym} and suppose that $W_K^\top W_Q$ is positive semi-definite. For a bandwidth $\e > 0$, let ${k^{\infty}_{\varepsilon}} = \sk({c}/{\e})$, that is the attention kernel for the Sinkformer with the cost ${c}/{\e}$. 
The mapping ${T}^{\infty}_{\mu, \e} : x \mapsto \frac{1}{\e}\int {k^{\infty}_{\varepsilon}}(x,x')W_V x'd\mu(x')$ corresponds to the continuous version of the Sinkformer where we re-scale $W_Q W_K^{T} = -W_V$ by ${\e}$. 
To better understand the dynamics of attention, we study the asymptotic regime in which the bandwidth $\varepsilon \to 0$.  In this regime, one can show that $\forall x \in \RR^d$, $\e T^{\infty}_{\mu,\e}(x) \to W_V x$ (details in Appendix \ref{app:proofs}). Thus, to go beyond first order, we study the modified map $\overline{T}^{\infty}_{\mu, \e} = {T}^{\infty}_{\mu, \e} - \frac{1}{\e} W_V.$ A natural question is the limit of this quantity when $\e \to 0$, and what the PDE defined by this limit is. We have the following theorem.
\begin{thm}[Sinkformer's PDE]
\label{thm:diffusion}
Let $\mu \in \Mm(\RR^d)$. Suppose that $\mu$ is supported on a compact set and has a density $\rho \in \Cc^{3}(\RR^d)$.  Suppose assumption \ref{asp:sym} and that  $W_K^\top W_Q$ is positive semi-definite. Then one has in $L^2$ norm as $\e \to 0$,
$$\overline{T}^{\infty}_{\mu, \e} \to \overline{T}^{\infty}_{\mu, 0} \coloneqq - \frac{\nabla \rho}{\rho}.$$
In this limit, the PDE $\partial_t\rho + \mathrm{div}(\rho \overline{T}^{\infty}_{\mu, 0}) = 0$ rewrites
\begin{equation}\label{eq:_laplacian_sink}
\partial_t \rho = \Delta \rho.
\end{equation}
\end{thm}
A proof is available in Appendix \ref{app:proofs}, making use of Theorem 1 from \citet{marshall2019manifold}.
We recover in Equation \eqref{eq:_laplacian_sink} the well-known \textbf{heat equation}. 

We want to compare this result with the one obtained with the SoftMax normalization. In order to carry a similar analysis, we make use of a Laplace expansion result \citep{tierney1989fully, singer2006graph}. However, the kernel ${k}^{1}_{\e} = \sm(c/\e)$ is not suited for using Laplace method because it does not always have a limit when $\e \to 0$. Thus, we consider the modified cost as in Proposition~\ref{prop:modularity}, $\tilde{c}(x, x') = -\frac{\|W_Qx - W_K x'\|^{2}}2$. The kernel $\tilde{k}^{1}_{\e} =\sm(\tilde{c}/\varepsilon)$, for which we can now apply Laplace expansion result, then corresponds to the L2 self-attention formulation \citep{kim2021lipschitz}. Note that thanks to Proposition \ref{prop:modularity}, $\tilde{k}^{\infty}_{\e} = {k}^{\infty}_{\e}$: Sinkorn's algorithm will have the same output for both costs. To simplify the expressions derived, we assume that $W_Q$ and $W_K$ are in $\RR^{d\times d}$ and are invertible. Similarly to the analysis conducted for Sinkformers, we consider the mapping ${T}^{1}_{\mu, \e} : x \mapsto \frac{1}{\e}\int {\tilde{k}^{1}_{\varepsilon}}(x,x')W_V x'd\mu(x')$. When $\e \to 0$, we show that $\forall x \in \RR^d$, $\e T^{1}_{\mu,\e}(x) \to - W_Q^{\top} W_Q x$ (details in Appendix \ref{app:proofs}). Thus, we consider $\overline{T}^{1}_{\mu, \e} = {T}^{1}_{\mu, \e} + \frac{1}{\e}\mathrm{W_Q^{\top} W_Q}$. We have the following result.

\begin{prop}[Transformer's PDE]
\label{prop:soft}
Let $\mu \in \Mm(\RR^d)$. Suppose that $\mu$ is supported on a compact set and has a density $\rho \in \Cc^{1}(\RR^d)$. Suppose assumption \ref{asp:sym} and that $W_Q$ and $W_K$ are in $\RR^{d\times d}$ and are invertible. Then one has $\forall x \in \RR^d$,
$$\overline{T}^{1}_{\mu, \e}(x) \to \overline{T}^{1}_{\mu, 0}(x) \coloneqq - W^{\top}_Q W^{-1} _K \frac{\nabla \rho}{\rho}(W^{-1}_K W_Q x ).$$
In this limit, the PDE $\partial_t\rho+ \mathrm{div}(\rho \overline{T}^{1}_{\mu, 0}) = 0$ rewrites
\begin{equation}\label{eq:_laplacian_softmax}
\partial_t \rho = \mathrm{div}(W^{\top}_Q W^{-1} _K \frac{\nabla \rho}{\rho}(W^{-1}_K W_Q \cdot ) \rho)\end{equation}
\end{prop}

A proof is given in Appendix \ref{app:proofs}.
While equation \eqref{eq:_laplacian_sink} corresponds to the heat equation, equation \eqref{eq:_laplacian_softmax} is different. First, it is nonlinear in $\rho$. Second, it is nonlocal since the evolution of the density at $x$ depends on the value of this density at location $W^{-1}_K W_Qx$. Note that the linear and local aspect of Sinkformer's PDE on the one hand, and the nonlinear and nonlocal aspect of Transformer's PDE on the other hand, remain true without assuming ${W_Q}^\top W_K = -W_V$ (details in Appendix \ref{app:proofs}).

\section{Experiments}\label{sec:experiments}
We now demonstrate the applicability of Sinkformers on a large variety of experiments with different modalities. We use Pytorch \citep{paszke2017automatic} and Nvidia Tesla V100 GPUs. Our code is open-sourced and is available at this address: \url{https://github.com/michaelsdr/sinkformers}. All the experimental details are given in Appendix \ref{app:exp_details}.

\paragraph{Practical implementation.}
In all our experiments, we use existing Transformer architectures and modify the SoftMax operator in attention modules with Sinkhorn's algorithm, which we implement in $\log$ domain for stability (details in Appendix \ref{app:imp_details}).

\subsection{ModelNet 40 classification}\label{sec:modelnet}

The ModelNet 40 dataset \citep{wu20153d} is composed of 40 popular object categories in 3D.
Transformers for point clouds and sets have been applied to the ModelNet 40 classification in several works, such as Set Transformers \citep{lee2019set} or Point Cloud Transformers \citep{guo2021pct}.

\paragraph{Set Sinkformers.} Set Transformers \citep{lee2019set} also have an encoder decoder structure with different possibilities for defining attention-based set operations. We propose to focus on the architecture that uses \textit{Induced Self Attention Block} (ISAB), which bypasses the quadratic time complexity of Self Attention Blocks (SAB). More details about this architecture can be found in \citep{lee2019set}. We reproduce the ModelNet 40 classification experiment using $5000$ uniformly sampled points for each shape and use a Set Transformer and a Set Sinkformer with two ISAB layers in the encoder and a decoder composed of a SAB and a Pooling by Multihead
Attention (PMA) module. While the reported test accuracy is of $87.8 \%$ using a Set Transformer, we obtain as our best accuracy when performing $21$ iterations of Sinkhorn algorithm within our Sinkformer of $89.1 \%$. Results are summarized in Table \ref{tab:results_MODELNET}.
\begin{figure}[H]
\includegraphics[width=\columnwidth]{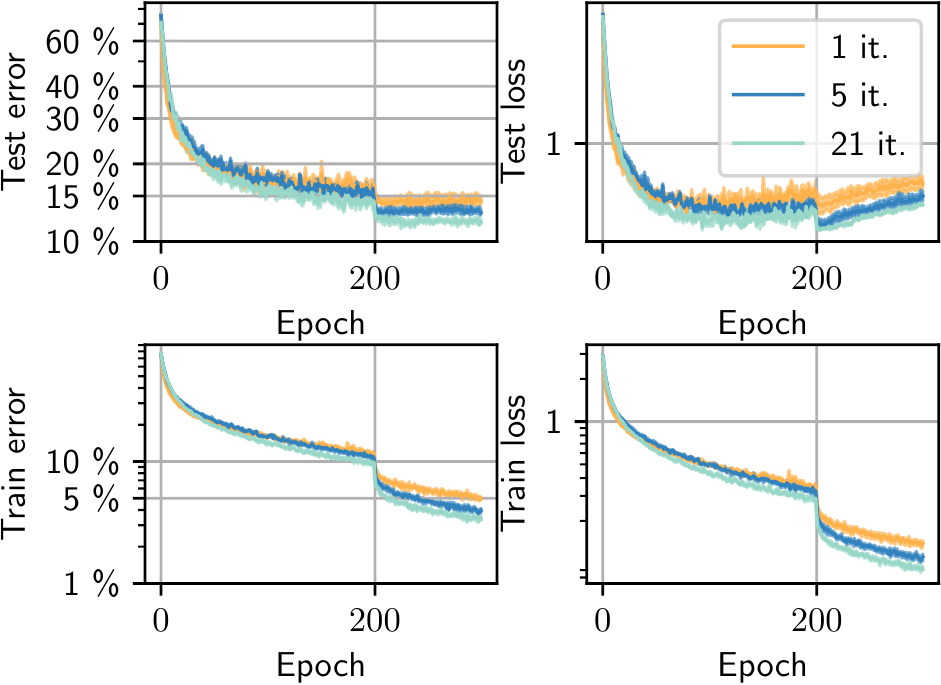} 
\caption{\textbf{Classification error and loss on ModelNet 40} when training a Set Transformer and a Set Sinkformer with different number of iterations in Sinkhorn's algorithm.}\label{fig:model_net_lc}
\vspace{-1em}
\end{figure}
\begin{figure*}[h]
 \centering
 \includegraphics[width=\textwidth]{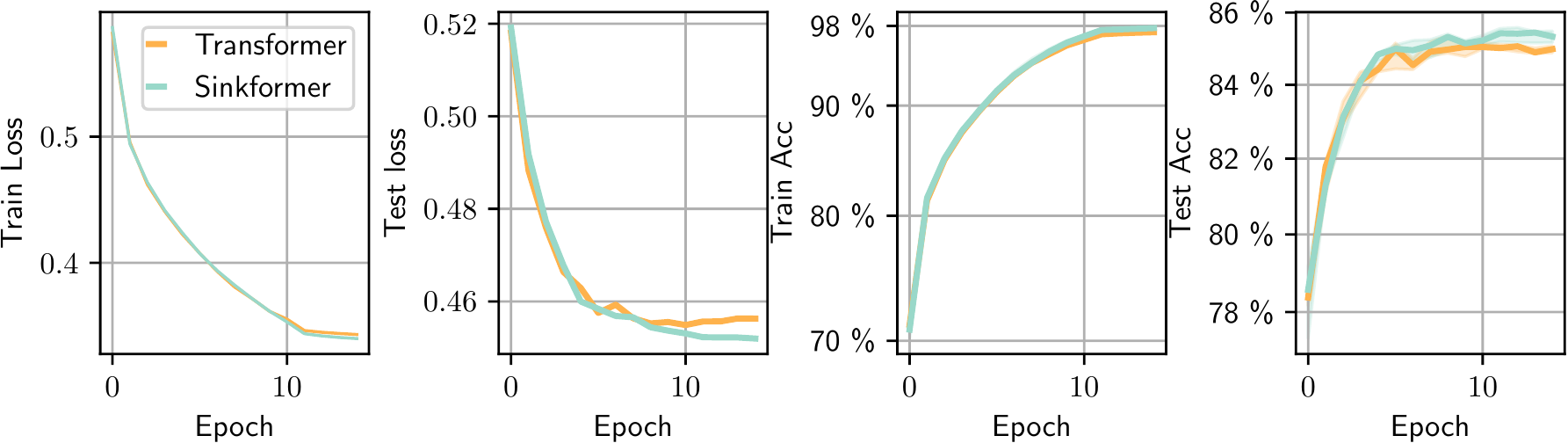} 
 \vspace{-2em}
 \caption{\textbf{Learning curves} when training a Transformer and a Sinkformer on the Sentiment Analysis task on the IMDb Dataset.}
\label{fig:sentiment_imbd}
\vspace{-1em}
 \end{figure*}
Moreover, we show in Figure \ref{fig:model_net_lc} the learning curves corresponding to this experiment. Interestingly, the number of iterations within Sinkhorn's algorithm increases the accuracy of the model. Note that we only consider an odd number of iterations since we always want to have row-wise stochastic attention matrices to be consistent with the properties of the SoftMax.

\paragraph{Point Cloud Transformers.} 

We also train Point Cloud Transformers \citep{guo2021pct} on ModelNet 40. This architecture achieves accuracy comparable to the state of the art on this dataset. We compare best and median test accuracy over $4$ runs. Results are reported in Table \ref{tab:results_MODELNET}, where we see that while the best test-accuracy is narrowly achieved for the Transformer, the Sinkformer has a slightly better median accuracy.

\begin{table}[h]
\vskip -0.15in
\centering
\caption{\label{tab:results_MODELNET}\textbf{Test accuracy for ModelNet 40} over 4 runs for each model.}
\vskip 0.15in
\begin{adjustbox}{width=\columnwidth,center}
\begin{tabular}{|l|l|l|l|l|}
  \hline
  \textbf{Model} &  \textbf{Best} &  \textbf{Median} & \textbf{Mean} & \textbf{Worst} \\ \Xhline{5\arrayrulewidth}
  {Set Transformer} & {$87.8 \%$}   & {$86.3 \%$} & {$85.8 \%$}   & {$84.7 \%$} \\ \hline
  {Set Sinkformer} & {$\mathbf{89.1} \%$}   & {$\mathbf{88.4} \%$} & {$\mathbf{88.3} \%$}   & {$\mathbf{88.1} \%$} \\ \Xhline{3\arrayrulewidth}
  {Point Cloud Transformer} & {$\mathbf{93.2} \%$}   & {$92.5 \%$} & {${92.5} \%$}   & {$92.3 \%$} \\ \hline
  {Point Cloud Sinkformer} & {$93.1 \%$}   & {$\mathbf{92.8} \%$} & {$\mathbf{92.7} \%$} & {$\mathbf{92.5} \%$} \\ \hline
\end{tabular}
\end{adjustbox}
\end{table}

\subsection{Sentiment Analysis}

We train a Transformer (composed of an attention-based encoder followed by a max-pooling layer) and a Sinkformer on the IMDb movie review dataset \citep{maas-EtAl:2011:ACL-HLT2011} for sentiment analysis. This text classification task consists of predicting whether a movie review is positive or negative. The learning curves are shown in Figure \ref{fig:sentiment_imbd}, with a gain in accuracy when using a Sinkformer. In this experiment, Sinkhorn's algorithm converges perfectly in 3 iterations (the resulting attention matrices are doubly stochastic), which corresponds to the green curve. The Sinkformer only adds a small computational overhead, since the training time per epoch is $4$m $02$s for the Transformer against $4$m $22$s for the Sinkformer. 

\subsection{Neural Machine Translation}\label{sec:fair}

We train a Transformer and its Sinkformer counterpart using the \texttt{fairseq} \citep{ott2019fairseq} sequence modeling toolkit on the IWSLT'14 German to English dataset \citep{cettolo2014report}. The architecture used is composed of an encoder and a decoder, both of depth $6$. We plug Sinkhorn's algorithm only into the encoder part. \textcolor{black}{Indeed, in the decoder, we can only pay attention to previous positions in the output sequence. For this reason, we need a mask that prevents a straightforward application of Sinkhorn's algorithm.} We demonstrate that even when using the hyper-parameters used to optimally train the Transformer, we achieve a similar BLEU~\citep{papineni2002bleu} over $6$ runs. We first train a Transformer for $30$ epochs. On the evaluation set, we obtain a BLEU of $34.43$. We then consider a Sinkformer with the weights of the trained Transformer. Interestingly, even this un-adapted Sinkformer provides a median BLEU score of $33.81$. We then divide the learning rate by $10$ and retrain for $5$ additional epochs both the Transformer and the Sinkformer to obtain a median BLEU of respectively $34.68$ and $34.73$ (Table \ref{tab:nmt}). Importantly, the runtime for one training epoch is almost the same for both models: $2$m $48$s (Transformer) against $2$m $52$s (Sinkformer).

\begin{table}[h]
\vskip -0.15in
\centering
\caption{\label{tab:results_CIFAR}\textbf{Median BLEU score} over 6 runs on the IWSLT'14 German to English dataset. The score ${}^{\mathbf {\star}}$ is when evaluating the Sinkformer with the weights of the trained Transformer.}
\vskip 0.15in
\begin{adjustbox}{width=0.8\columnwidth,center}
\begin{tabular}{|l|l|l|}
  \hline
  \textbf{Model} &  Epoch 30 &  Epoch 35 \\ \Xhline{3\arrayrulewidth}
  {Transformer} & {$34.43$}   & {$34.68$} \\ \hline
  {Sinkformer} & {$33.81^{\mathbf {\star}}$}   & {$\mathbf{34.73}$} \\ \hline
\end{tabular}\label{tab:nmt}
\end{adjustbox}
\end{table}

\subsection{Vision Transformers}\label{sec:vit}

Vision Transformers (ViT) \citep{dosovitskiy2020image} have recently emerged as a promising architecture for achieving state of the art performance on computer vision tasks \citep{zhai2021scaling}, using only attention based mechanisms by selecting patches of fixed size in images and feeding them into an attention mechanism. 
\begin{figure}[H]
    \begin{minipage}[c]{0.56\linewidth}
        \includegraphics[width=\textwidth]{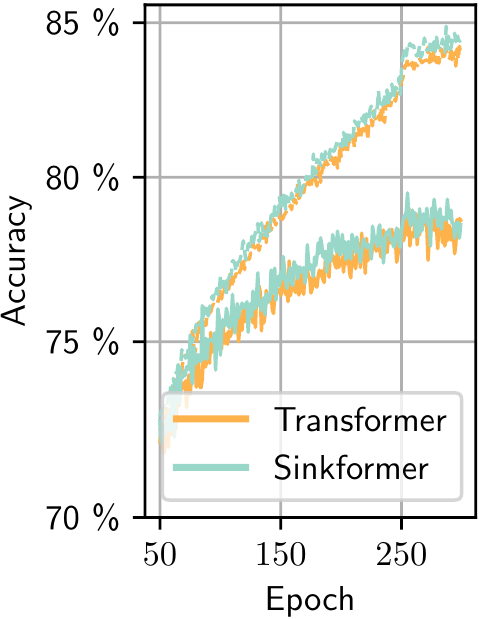}
    \end{minipage}\hfill
    \begin{minipage}[c]{0.37\linewidth}
    \caption{\textbf{Train} (dotted) and \textbf{test} (plain) \textbf{accuracy} as a function of the number of epochs when training a ViT and its Sinkformer counterpart on the cats and dogs classification task (median over 5 runs). }\label{fig:vit}
    \end{minipage}
\end{figure}
\paragraph{Cats and dogs classification.}
We train a ViT and its Sinkformer counterpart on a binary cats and dogs image classification task. 
The evolution of the train and test accuracy is displayed in Figure \ref{fig:vit}. The \textit{median} test accuracy is $79.0 \%$ for the Transformer against $79.5\%$ for the Sinkformer, whereas the \textit{maximum} test accuracy is $80.0 \%$ for the Transformer against $80.5\%$ for the Sinkformer. \textcolor{black}{We also use $3$ iterations in Sinkhorn's algorithm which leads to a negligible computational overhead (training time per epoch of 3m 25s for the Sinkformer against 3m 20s for the Transformer).}

\paragraph{Impact of the patch size on the final accuracy.}

\begin{figure}[h]
    \begin{minipage}[c]{0.50\linewidth}
        \includegraphics[width=\textwidth]{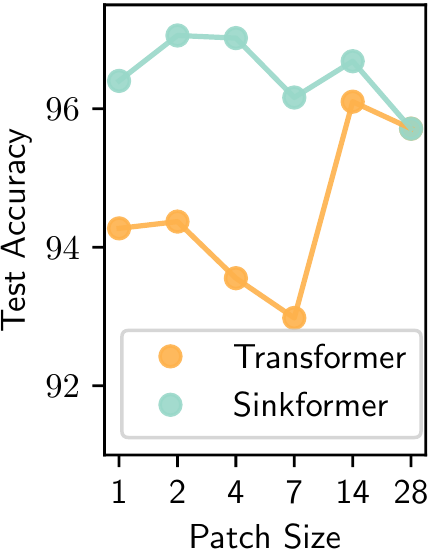}
    \end{minipage}\hfill
    \begin{minipage}[c]{0.4\linewidth}
    \caption{\textbf{Final test accuracy} when training a one layer and one head self attention module on the MNIST dataset, with no feedforward neural network, when varying the patch size (median over 5 runs).}\label{fig:patches}
    \end{minipage}
\end{figure}

We consider a one-layer and one-head self-attention module on the MNIST dataset, with no additional layer. The purpose is to isolate the self-attention module and study how its accuracy is affected by the choice of the patch size. Results are displayed in Figure \ref{fig:patches}. We recall that a MNIST image is of size $28 \times 28$. When taking only one patch of size $28$, both models are equivalent because the attention matrix is of size $1$. However, when the patch size gets smaller, the two models are different and the Sinkformer outperforms the Transformer.

\vspace{-0.3cm}
\section*{Conclusion}

In this paper, we presented the Sinkformer, a variant of the Transformer in which the SoftMax, which leads to row-wise stochastic attention, is replaced by Sinkhorn's algorithm, which leads to doubly stochastic attention. This new model is motivated by the  empirical finding that attention matrices in Transformers get closer and closer to doubly stochastic matrices during the training process. This modification is easily implemented in practice by simply replacing the SoftMax in the attention modules of existing Transformers without changing any parameter in the network. It also provides a new framework for theoretically studying attention-based mechanisms, such as the interpretation of Sinkformers as Wasserstein gradient flows in the infinitesimal step size regime or as diffusion operators in the mean-field limit. On the experimental side, Sinkformers lead to better accuracy in a variety of experiments: classification of 3D shapes, sentiment analysis, neural machine translation, and image classification.
\section*{Acknowledgments}

This work was granted access to the HPC resources of IDRIS under the allocation 2020-[AD011012073] made by GENCI. This work was supported in part by the French government under management of Agence Nationale de la Recherche as part of the “Investissements d’avenir” program, reference ANR19-P3IA-0001 (PRAIRIE 3IA Institute). This work was
supported in part by the European Research Council (ERC project NORIA).  We thank Marco Cuturi and D. Sculley for their comments on a draft of the paper. We thank Scott Pesme, Pierre Rizkallah, Othmane Sebbouh, Thibault Séjourné and the anonymous reviewers for helpful feedbacks. 
\bibliography{references}

\begin{thebibliography}{}

\bibitem[Agrawal et~al., 2019]{agrawal2019differentiable}
Agrawal, A., Amos, B., Barratt, S., Boyd, S., Diamond, S., and Kolter, Z.
  (2019).
\newblock Differentiable convex optimization layers.
\newblock {\em arXiv preprint arXiv:1910.12430}.

\bibitem[Bahdanau et~al., 2014]{bahdanau_2014}
Bahdanau, D., Cho, K., and Bengio, Y. (2014).
\newblock Neural machine translation by jointly learning to align and
  translate.
\newblock {\em arXiv preprint arXiv:1409.0473}.

\bibitem[Bai et~al., 2019]{bai2019deep}
Bai, S., Kolter, J.~Z., and Koltun, V. (2019).
\newblock Deep equilibrium models.
\newblock {\em Advances in Neural Information Processing Systems}, 32:690--701.

\bibitem[Behrmann et~al., 2019]{behrmann2019invertible}
Behrmann, J., Grathwohl, W., Chen, R.~T., Duvenaud, D., and Jacobsen, J.-H.
  (2019).
\newblock Invertible residual networks.
\newblock In {\em International Conference on Machine Learning}, pages
  573--582. PMLR.

\bibitem[Blondel et~al., 2021]{blondel2021efficient}
Blondel, M., Berthet, Q., Cuturi, M., Frostig, R., Hoyer, S.,
  Llinares-L{\'o}pez, F., Pedregosa, F., and Vert, J.-P. (2021).
\newblock Efficient and modular implicit differentiation.
\newblock {\em arXiv preprint arXiv:2105.15183}.

\bibitem[Brown et~al., 2020]{brown2020language}
Brown, T.~B., Mann, B., Ryder, N., Subbiah, M., Kaplan, J., Dhariwal, P.,
  Neelakantan, A., Shyam, P., Sastry, G., Askell, A., et~al. (2020).
\newblock Language models are few-shot learners.
\newblock {\em arXiv preprint arXiv:2005.14165}.

\bibitem[Cettolo et~al., 2014]{cettolo2014report}
Cettolo, M., Niehues, J., St{\"u}ker, S., Bentivogli, L., and Federico, M.
  (2014).
\newblock Report on the 11th iwslt evaluation campaign, iwslt 2014.
\newblock In {\em Proceedings of the International Workshop on Spoken Language
  Translation, Hanoi, Vietnam}, volume~57.

\bibitem[Charlier et~al., 2021]{charlier2021kernel}
Charlier, B., Feydy, J., Glaun{\`e}s, J., Collin, F.-D., and Durif, G. (2021).
\newblock Kernel operations on the gpu, with autodiff, without memory
  overflows.
\newblock {\em Journal of Machine Learning Research}, 22(74):1--6.

\bibitem[Chen et~al., 2018]{chen2018neural}
Chen, T.~Q., Rubanova, Y., Bettencourt, J., and Duvenaud, D.~K. (2018).
\newblock Neural ordinary differential equations.
\newblock In {\em Advances in neural information processing systems}, pages
  6571--6583.

\bibitem[Cuturi, 2013]{cuturi2013sinkhorn}
Cuturi, M. (2013).
\newblock Sinkhorn distances: Lightspeed computation of optimal transport.
\newblock {\em Advances in neural information processing systems},
  26:2292--2300.

\bibitem[Cuturi et~al., 2020]{cuturi_2020}
Cuturi, M., Teboul, O., Niles-Weed, J., and Vert, J.-P. (2020).
\newblock Supervised quantile normalization for low rank matrix factorization.
\newblock In {\em International Conference on Machine Learning}, pages
  2269--2279. PMLR.

\bibitem[De~Bie et~al., 2019]{de2019stochastic}
De~Bie, G., Peyr{\'e}, G., and Cuturi, M. (2019).
\newblock Stochastic deep networks.
\newblock In {\em International Conference on Machine Learning}, pages
  1556--1565. PMLR.

\bibitem[Dong et~al., 2021]{dong2021attention}
Dong, Y., Cordonnier, J.-B., and Loukas, A. (2021).
\newblock Attention is not all you need: Pure attention loses rank doubly
  exponentially with depth.
\newblock {\em arXiv preprint arXiv:2103.03404}.

\bibitem[Dosovitskiy et~al., 2020]{dosovitskiy2020image}
Dosovitskiy, A., Beyer, L., Kolesnikov, A., Weissenborn, D., Zhai, X.,
  Unterthiner, T., Dehghani, M., Minderer, M., Heigold, G., Gelly, S., et~al.
  (2020).
\newblock An image is worth 16x16 words: Transformers for image recognition at
  scale.
\newblock {\em arXiv preprint arXiv:2010.11929}.

\bibitem[Gould et~al., 2016]{gould2016differentiating}
Gould, S., Fernando, B., Cherian, A., Anderson, P., Cruz, R.~S., and Guo, E.
  (2016).
\newblock On differentiating parameterized argmin and argmax problems with
  application to bi-level optimization.
\newblock {\em arXiv preprint arXiv:1607.05447}.

\bibitem[Gould et~al., 2019]{gould2019deep}
Gould, S., Hartley, R., and Campbell, D. (2019).
\newblock Deep declarative networks: A new hope.
\newblock {\em arXiv preprint arXiv:1909.04866}.

\bibitem[Griewank and Walther, 2008]{griewank_2008}
Griewank, A. and Walther, A. (2008).
\newblock {\em Evaluating derivatives: principles and techniques of algorithmic
  differentiation}.
\newblock SIAM.

\bibitem[Guo et~al., 2021]{guo2021pct}
Guo, M.-H., Cai, J.-X., Liu, Z.-N., Mu, T.-J., Martin, R.~R., and Hu, S.-M.
  (2021).
\newblock Pct: Point cloud transformer.
\newblock {\em Computational Visual Media}, 7(2):187--199.

\bibitem[He et~al., 2016]{he2016deep}
He, K., Zhang, X., Ren, S., and Sun, J. (2016).
\newblock Deep residual learning for image recognition.
\newblock In {\em Proceedings of the IEEE conference on computer vision and
  pattern recognition}, pages 770--778.

\bibitem[Hein et~al., 2007]{hein2007graph}
Hein, M., Audibert, J.-Y., and Luxburg, U.~v. (2007).
\newblock Graph laplacians and their convergence on random neighborhood graphs.
\newblock {\em Journal of Machine Learning Research}, 8(6).

\bibitem[Jordan et~al., 1998]{jordan1998variational}
Jordan, R., Kinderlehrer, D., and Otto, F. (1998).
\newblock The variational formulation of the fokker--planck equation.
\newblock {\em SIAM journal on mathematical analysis}, 29(1):1--17.

\bibitem[Kim et~al., 2021]{kim2021lipschitz}
Kim, H., Papamakarios, G., and Mnih, A. (2021).
\newblock The lipschitz constant of self-attention.
\newblock In {\em International Conference on Machine Learning}, pages
  5562--5571. PMLR.

\bibitem[Kingma and Ba, 2014]{kingma2014adam}
Kingma, D.~P. and Ba, J. (2014).
\newblock Adam: A method for stochastic optimization.
\newblock {\em arXiv preprint arXiv:1412.6980}.

\bibitem[Krantz and Parks, 2012]{krantz_2012}
Krantz, S.~G. and Parks, H.~R. (2012).
\newblock {\em The implicit function theorem: history, theory, and
  applications}.
\newblock Springer Science \& Business Media.

\bibitem[Lee et~al., 2019]{lee2019set}
Lee, J., Lee, Y., Kim, J., Kosiorek, A., Choi, S., and Teh, Y.~W. (2019).
\newblock Set transformer: A framework for attention-based
  permutation-invariant neural networks.
\newblock In {\em International Conference on Machine Learning}, pages
  3744--3753. PMLR.

\bibitem[Lu et~al., 2018]{lu2017finite}
Lu, Y., Zhong, A., Li, Q., and Dong, B. (2018).
\newblock Beyond finite layer neural networks: Bridging deep architectures and
  numerical differential equations.
\newblock In {\em International Conference on Machine Learning}, pages
  3276--3285. PMLR.

\bibitem[Luise et~al., 2018]{luise_2018}
Luise, G., Rudi, A., Pontil, M., and Ciliberto, C. (2018).
\newblock Differential properties of sinkhorn approximation for learning with
  wasserstein distance.
\newblock {\em arXiv preprint arXiv:1805.11897}.

\bibitem[Maas et~al., 2011]{maas-EtAl:2011:ACL-HLT2011}
Maas, A.~L., Daly, R.~E., Pham, P.~T., Huang, D., Ng, A.~Y., and Potts, C.
  (2011).
\newblock Learning word vectors for sentiment analysis.
\newblock In {\em Proceedings of the 49th Annual Meeting of the Association for
  Computational Linguistics: Human Language Technologies}, pages 142--150,
  Portland, Oregon, USA. Association for Computational Linguistics.

\bibitem[Marshall and Coifman, 2019]{marshall2019manifold}
Marshall, N.~F. and Coifman, R.~R. (2019).
\newblock Manifold learning with bi-stochastic kernels.
\newblock {\em IMA Journal of Applied Mathematics}, 84(3):455--482.

\bibitem[Mialon et~al., 2021]{mialon2021trainable}
Mialon, G., Chen, D., d'Aspremont, A., and Mairal, J. (2021).
\newblock A trainable optimal transport embedding for feature aggregation and
  its relationship to attention.
\newblock In {\em ICLR 2021-The Ninth International Conference on Learning
  Representations}.

\bibitem[Milanfar, 2013]{milanfar2013symmetrizing}
Milanfar, P. (2013).
\newblock Symmetrizing smoothing filters.
\newblock {\em SIAM Journal on Imaging Sciences}, 6(1):263--284.

\bibitem[Niculae et~al., 2018]{niculae_2018}
Niculae, V., Martins, A., Blondel, M., and Cardie, C. (2018).
\newblock Sparsemap: Differentiable sparse structured inference.
\newblock In {\em International Conference on Machine Learning}, pages
  3799--3808. PMLR.

\bibitem[Ott et~al., 2019]{ott2019fairseq}
Ott, M., Edunov, S., Baevski, A., Fan, A., Gross, S., Ng, N., Grangier, D., and
  Auli, M. (2019).
\newblock fairseq: A fast, extensible toolkit for sequence modeling.
\newblock In {\em Proceedings of NAACL-HLT 2019: Demonstrations}.

\bibitem[Papineni et~al., 2002]{papineni2002bleu}
Papineni, K., Roukos, S., Ward, T., and Zhu, W.-J. (2002).
\newblock Bleu: a method for automatic evaluation of machine translation.
\newblock In {\em Proceedings of the 40th annual meeting of the Association for
  Computational Linguistics}, pages 311--318.

\bibitem[Paszke et~al., 2017]{paszke2017automatic}
Paszke, A., Gross, S., Chintala, S., Chanan, G., Yang, E., DeVito, Z., Lin, Z.,
  Desmaison, A., Antiga, L., and Lerer, A. (2017).
\newblock Automatic differentiation in pytorch.

\bibitem[Peng et~al., 2017]{peng2017large}
Peng, C., Zhang, X., Yu, G., Luo, G., and Sun, J. (2017).
\newblock Large kernel matters--improve semantic segmentation by global
  convolutional network.
\newblock In {\em Proceedings of the IEEE conference on computer vision and
  pattern recognition}, pages 4353--4361.

\bibitem[Peyr{\'e} et~al., 2019]{peyre2019computational}
Peyr{\'e}, G., Cuturi, M., et~al. (2019).
\newblock Computational optimal transport: With applications to data science.
\newblock {\em Foundations and Trends{\textregistered} in Machine Learning},
  11(5-6):355--607.

\bibitem[Radford et~al., 2019]{radford2019language}
Radford, A., Wu, J., Child, R., Luan, D., Amodei, D., Sutskever, I., et~al.
  (2019).
\newblock Language models are unsupervised multitask learners.
\newblock {\em OpenAI blog}, 1(8):9.

\bibitem[Renardy and Rogers, 2006]{renardy_2006}
Renardy, M. and Rogers, R.~C. (2006).
\newblock {\em An introduction to partial differential equations}, volume~13.
\newblock Springer Science \& Business Media.

\bibitem[Ruder, 2016]{ruder2016overview}
Ruder, S. (2016).
\newblock An overview of gradient descent optimization algorithms.
\newblock {\em arXiv preprint arXiv:1609.04747}.

\bibitem[Ruthotto and Haber, 2019]{ruthotto2018deep}
Ruthotto, L. and Haber, E. (2019).
\newblock Deep neural networks motivated by partial differential equations.
\newblock {\em Journal of Mathematical Imaging and Vision}, pages 1--13.

\bibitem[Sander et~al., 2021]{pmlr-v139-sander21a}
Sander, M.~E., Ablin, P., Blondel, M., and Peyr{\'e}, G. (2021).
\newblock Momentum residual neural networks.
\newblock In {\em Proceedings of the 38th International Conference on Machine
  Learning}, volume 139 of {\em Proceedings of Machine Learning Research},
  pages 9276--9287. PMLR.

\bibitem[Santambrogio, 2017]{santambrogio2017euclidean}
Santambrogio, F. (2017).
\newblock $\{$Euclidean, metric, and Wasserstein$\}$ gradient flows: an
  overview.
\newblock {\em Bulletin of Mathematical Sciences}, 7(1):87--154.

\bibitem[Singer, 2006]{singer2006graph}
Singer, A. (2006).
\newblock From graph to manifold laplacian: The convergence rate.
\newblock {\em Applied and Computational Harmonic Analysis}, 21(1):128--134.

\bibitem[Sinkhorn, 1964]{sinkhorn1964relationship}
Sinkhorn, R. (1964).
\newblock A relationship between arbitrary positive matrices and doubly
  stochastic matrices.
\newblock {\em The annals of mathematical statistics}, 35(2):876--879.

\bibitem[Song et~al., 2018]{song2018mean}
Song, M., Montanari, A., and Nguyen, P. (2018).
\newblock A mean field view of the landscape of two-layers neural networks.
\newblock {\em Proceedings of the National Academy of Sciences},
  115:E7665--E7671.

\bibitem[Sun et~al., 2018]{sun2018stochastic}
Sun, Q., Tao, Y., and Du, Q. (2018).
\newblock Stochastic training of residual networks: a differential equation
  viewpoint.
\newblock {\em arXiv preprint arXiv:1812.00174}.

\bibitem[Tay et~al., 2020]{tay2020sparse}
Tay, Y., Bahri, D., Yang, L., Metzler, D., and Juan, D.-C. (2020).
\newblock Sparse sinkhorn attention.
\newblock In {\em International Conference on Machine Learning}, pages
  9438--9447. PMLR.

\bibitem[Teh et~al., 2019]{dupont2019augmented}
Teh, Y., Doucet, A., and Dupont, E. (2019).
\newblock Augmented neural odes.
\newblock {\em Advances in Neural Information Processing Systems 32 (NIPS
  2019)}, 32(2019).

\bibitem[Tierney et~al., 1989]{tierney1989fully}
Tierney, L., Kass, R.~E., and Kadane, J.~B. (1989).
\newblock Fully exponential laplace approximations to expectations and
  variances of nonpositive functions.
\newblock {\em Journal of the american statistical association},
  84(407):710--716.

\bibitem[Ting et~al., 2011]{ting2011analysis}
Ting, D., Huang, L., and Jordan, M. (2011).
\newblock An analysis of the convergence of graph laplacians.
\newblock {\em arXiv preprint arXiv:1101.5435}.

\bibitem[Vaswani et~al., 2017]{vaswani2017attention}
Vaswani, A., Shazeer, N., Parmar, N., Uszkoreit, J., Jones, L., Gomez, A.~N.,
  Kaiser, {\L}., and Polosukhin, I. (2017).
\newblock Attention is all you need.
\newblock In {\em Advances in neural information processing systems}, pages
  5998--6008.

\bibitem[Vuckovic et~al., 2021]{vuckovic2021regularity}
Vuckovic, J., Baratin, A., and Combes, R. T.~d. (2021).
\newblock On the regularity of attention.
\newblock {\em arXiv preprint arXiv:2102.05628}.

\bibitem[Wang et~al., 2018]{wang2018superneurons}
Wang, L., Ye, J., Zhao, Y., Wu, W., Li, A., Song, S.~L., Xu, Z., and Kraska, T.
  (2018).
\newblock Superneurons: Dynamic gpu memory management for training deep neural
  networks.
\newblock In {\em Proceedings of the 23rd ACM SIGPLAN Symposium on Principles
  and Practice of Parallel Programming}, pages 41--53.

\bibitem[Weinan, 2017]{E_2017}
Weinan, E. (2017).
\newblock A proposal on machine learning via dynamical systems.
\newblock {\em Communications in Mathematics and Statistics}, 5(1):1--11.

\bibitem[Weinan et~al., 2019]{E_2018}
Weinan, E., Han, J., and Li, Q. (2019).
\newblock A mean-field optimal control formulation of deep learning.
\newblock {\em Research in the Mathematical Sciences}, 6(1):10.

\bibitem[Wolf et~al., 2019]{wolf2019huggingface}
Wolf, T., Debut, L., Sanh, V., Chaumond, J., Delangue, C., Moi, A., Cistac, P.,
  Rault, T., Louf, R., Funtowicz, M., et~al. (2019).
\newblock Huggingface's transformers: State-of-the-art natural language
  processing.
\newblock {\em arXiv preprint arXiv:1910.03771}.

\bibitem[Wormell and Reich, 2021]{wormell2021spectral}
Wormell, C.~L. and Reich, S. (2021).
\newblock Spectral convergence of diffusion maps: Improved error bounds and an
  alternative normalization.
\newblock {\em SIAM Journal on Numerical Analysis}, 59(3):1687--1734.

\bibitem[Wu et~al., 2015]{wu20153d}
Wu, Z., Song, S., Khosla, A., Yu, F., Zhang, L., Tang, X., and Xiao, J. (2015).
\newblock 3d shapenets: A deep representation for volumetric shapes.
\newblock In {\em Proceedings of the IEEE conference on computer vision and
  pattern recognition}, pages 1912--1920.

\bibitem[Yun et~al., 2019]{yun2019transformers}
Yun, C., Bhojanapalli, S., Rawat, A.~S., Reddi, S.~J., and Kumar, S. (2019).
\newblock Are transformers universal approximators of sequence-to-sequence
  functions?
\newblock {\em arXiv preprint arXiv:1912.10077}.

\bibitem[Zhai et~al., 2021]{zhai2021scaling}
Zhai, X., Kolesnikov, A., Houlsby, N., and Beyer, L. (2021).
\newblock Scaling vision transformers.
\newblock {\em arXiv preprint arXiv:2106.04560}.

\bibitem[Zhao et~al., 2020]{zhao2020point}
Zhao, H., Jiang, L., Jia, J., Torr, P., and Koltun, V. (2020).
\newblock Point transformer.
\newblock {\em arXiv preprint arXiv:2012.09164}.

\bibitem[Zhu et~al., 2017]{zhu2017unpaired}
Zhu, J.-Y., Park, T., Isola, P., and Efros, A.~A. (2017).
\newblock Unpaired image-to-image translation using cycle-consistent
  adversarial networks.
\newblock In {\em Proceedings of the IEEE international conference on computer
  vision}, pages 2223--2232.

\bibitem[Zweig and Bruna, 2021]{zweig2021functional}
Zweig, A. and Bruna, J. (2021).
\newblock A functional perspective on learning symmetric functions with neural
  networks.
\newblock In {\em International Conference on Machine Learning}, pages
  13023--13032. PMLR.

\end{thebibliography}

\appendix
\onecolumn
\section*{Appendix}
In Section~\ref{app:proofs} we give the proofs of all the Propositions and the Theorem.  %
In Section~\ref{app:imp_details} we present the implementation details of Sinkformers. Section~\ref{app:exp_details} gives details for the experiments in the paper.

\section{Proofs}\label{app:proofs}

\subsection{Invariance  to  the  cost  function - Proof of Proposition \ref{prop:modularity}}
\begin{proof}
We use the variational formulation for Sinkhorn \citep{peyre2019computational}:
$$\sk(C) = \argmin_{K\mathbb{1} _{n} = K^{\top}\mathbb{1} _{n} = \mathbb{1} _{n}} \mathrm{KL}(K|K^{0})$$ with  $$\mathrm{KL}(K|K^{0}) = \sum_{i,j}K_{i,j}\log(\frac{K_{i,j}}{K^0_{i,j}}),$$
where $K^0_{i,j} = \exp(C_{i,j})$. 

We let $\tilde{C}_{i,j} = C_{ij} + f_i + g_j$. 
We have for $K \in U \coloneqq \{ K | K\mathbb{1} _{n} = K^{\top}\mathbb{1} _{n} = \mathbb{1} _{n}\}$ that
$
\mathrm{KL}(K|e^{\tilde{C}}) = \sum_{i,j}K_{i,j}\log(\frac{K_{i,j}}{e^{C_{i,j} + f_i + g_j}}) 
$. This gives 
$$
\mathrm{KL}(K|e^{\tilde{C}}) = \sum_{i,j}K_{i,j}[\log(\frac{K_{i,j}}{e^{C_{i,j}}}) - f_i - g_j] = \sum_{i,j}K_{i,j}[\log(\frac{K_{i,j}}{e^{C_{i,j}}})] - \sum_{i} f_i - \sum_{j} g_j$$
so that 
$$
\mathrm{KL}(K|e^{\tilde{C}}) = \mathrm{KL}(K|e^{{C}}) - \sum_{i} f_i - \sum_{j} g_j.
$$ 
This shows that $\mathrm{KL}(K|e^{\tilde{C}})$ and $\mathrm{KL}(K|e^{{C}})$ have the same argmin on $U$ which implies that $\sk(C) = \sk(\tilde{C})$.

\end{proof}
\subsection{PDEs associated with $\mathbf{k^0, k^1, k^{\infty}}$ - Proof of Proposition \ref{prop:gradient_flows}}

\begin{proof}
 Recall that for $p \in \{0, 1, \infty \}$, we have $T_{\mu}^{p}(x) = \int k^{p}(x, x') W_V x' d \mu(x')$. 

For $h \in \Cc(\RR^d \times \RR^d)$
consider 
$$
\Hh(\mu) = \int h(x,y) d\mu(x) d\mu(y).
$$
Then we have \citep{santambrogio2017euclidean}
$$
\frac{\delta \Hh}{\delta \mu}(\mu) = \int( h(x,.) + h(.,x))d\mu(x).
$$
We can now derive the different gradient expressions for $T_{\mu}^{0}$, $T_{\mu}^{1}$ and $T_{\mu}^{\infty}$.

For $T_{\mu}^{0}$: under Assumption \ref{asp:sym}, we have that $f(x, x') = e^{C(x,x')}$ is symmetric. This gives 
$$
\frac{\delta \Ff}{\delta \mu}(\mu) = \int f(.,x') d\mu(x')
$$
and by differentiation under the integral, under sufficient regularity assumptions on $\mu$, this gives 
$$
\nabla_{W}(\Ff^0)(x) = \int \nabla_x f(x,x')d\mu(x') = \int f(x,x')\nabla_x c(x,x')d\mu(x').
$$
Since $\nabla_x c (x,x') = -W_Vx'$, we get 
$$
\nabla_{W}(\Ff^0)(x) = - \int e^{c(x,x')}W_Vx' d\mu(x').
$$
For $\mu = \frac{1}{n} \sum_{i=1}^{n}\delta_{x_i}$ this is exactly 
$$
\nabla_{W}(\Ff^0)(x) = -\sum^{n}_{j=1}  K^{0}_{i,j}  W_V x_j.
$$
For $T_{\mu}^{1}$: we have
$$
\nabla[\log(\int e^{c(., x')} d \mu(x'))](x) = \int \frac{\nabla_x c (x, x') e^{c(x, x')}}{\int e^{c(x, y)} d \mu(y)} d \mu(x') = -\int \frac{ e^{c(x, x')} W_Vx'}{\int e^{c(x, y)} d \mu(y)} d \mu(x').
$$
For $T_{\mu}^{2}$: one has the dual formulation for $\Ff^{\infty}$ \citep{peyre2019computational}:

\begin{equation}\label{eq:entropic-ot}
    2 \Ff^{\infty}(\mu) = - \umax{f} \int_{\RR^d} (f +  f^c) \text{d} \mu
\end{equation}
where we denote the soft $c$ transform as
\begin{equation}\label{eq:c-transform}
    f^c(x') \coloneqq -\log\pa{\int e^{f(x)+c(x,x')} \text{d} \mu(x) }, 
\end{equation}
which actually depends on $\mu$ and $c$. 
One has for an optimal pair $f = f^c$ \citep{peyre2019computational}.
In addition, one has $k^{\infty}(x,x') = e^{c(x,x') +f(x) + f(x')}$. The Wasserstein gradient of $\Ff^{\infty}$ is then 
\begin{equation*}
    \nabla_W \Ff^{\infty}(\mu)  =  -\nabla f 
\end{equation*}
where $f$ is an optimal solution of~\eqref{eq:entropic-ot} (which is unique up to a constant).
The gradient of $f$ can be obtained using \eqref{eq:c-transform} and the fact that $f = f^c$: 
$$
\nabla f(x) = - \int e^{f(x) + f(x') + c(x,x')} \nabla_x c(x,x') d\mu(x') = -\int k^{\infty}(x,x') \nabla_x c(x,x') d\mu(x').
$$
This finally gives
\begin{equation}\label{eq:sinkhorn-map}
    \nabla_W \Ff^{\infty}(\mu) : x \mapsto
         -\int k^{\infty}(x,x') W_v x' \text{d} \mu(x'),
\end{equation}
that is what we wanted to show. 
\end{proof}
\subsection{The SoftMax normalization does not correspond to a gradient flow - Proof of Proposition \ref{prop:not_a_gradient}}
\begin{proof}
Suppose by contradiction that $T_{\mu}^{1} = -\nabla[\log(\int k^{0}(\cdot, x') d \mu(x'))]$ is a Wasserstein gradient. This implies that there exists a function $F$ such that, $\forall \mu \in \Mm(\RR^d)$ and $\forall x \in \RR^d$, 
$$
\frac{\delta F}{\delta \mu}(\mu)(x) = \log(\int k^{0}(\cdot, x') d \mu(x')).$$
We therefore have 
$$
\frac{\delta^2 F}{\delta \mu^2}(\mu)(x, x') = \frac{k^{0}(x,x')}{\int k^{0}(x, y) d \mu(y)},
$$
$\forall x, x' \in \RR^d$. However, $\frac{\delta^2 F}{\delta \mu^2}(\mu)$ is symmetric for all $\mu \in \Mm(\RR^d)$. The relationship $\frac{\delta^2 F}{\delta \mu^2}(\mu)(x, x') = \frac{\delta^2 F}{\delta \mu^2}(\mu)(x', x)$ then implies that for all $\mu$, $x$ and $x'$ such that $k^0(x, x') \neq 0$ we have
$$
\int k^{0}(x, y) d \mu(y) = \int k^{0}(x', y) d \mu(y).
$$
Taking $\mu = \delta_{y}$ gives $k^0(x, y) = k^0(x', y)$, which by symmetry implies that $k^0$ is a constant.

This is a contradiction since $k^0(x, x') = \exp(x^\top W_Q^\top W_Kx')$.
\end{proof}
\subsection{Sinkformer's PDE - Proof of Theorem \ref{thm:diffusion}}
\begin{proof}
Since $W_Q^\top W_K$ is positive-definite we write it $W_Q^\top W_K = A^2$ where $A$ is positive-definite. Note that thanks to Proposition \ref{prop:modularity}, if $\kappa_{\e}(x, x') = \exp(-\frac{\|x-x'\|^2}{2 \e})$, one has under Assumption \ref{asp:sym} that $\kappa^{\infty}_{\e}(Ax, Ax') = k^{\infty}_{\e}(x, x')$. For $x \in \RR^d$, we have 
$$\overline{T}^{\infty}_{\mu, \e}(A^{-1}x) = \frac{1}{\e} ( \int \kappa^{\infty}_{\e}(x, Ax')W_Vx'\rho(x')dx' - W_VA^{-1}x).$$
We perform the change of variable $y = Ax'$. This gives
$$
\overline{T}^{\infty}_{\mu, \e}(A^{-1}x) = \frac{1}{\e} ( \int \kappa^{\infty}_{\e}(x, y)W_VA^{-1} y\rho(A^{-1}y)C_Ady - W_VA^{-1}x),
$$
where $C_A$ depends only on $A$. We then apply Theorem 1 from \cite{marshall2019manifold} with $f = W_VA^{-1}$, $q(x) = \rho(A^{-1}x)$ and $w=\frac1{C_A}$, to obtain that 
$$
\overline{T}^{\infty}_{\mu, \e}(A^{-1}\cdot) \to \frac{2 \nabla f \nabla(q^{1/2})}{q^{1/2}} = W_VA^{-1} \frac{\nabla q}{q}
$$
in $L^2$ norm. Since $q(x) = \rho(A^{-1}x)$ we have obtained that $\frac{\nabla q}{q} = A^{-1} \frac {\nabla \rho}{\rho}(A^{-1} \cdot)$
so that 
$$
\overline{T}^{\infty}_{\mu, \e}(A^{-1}x) \to W_VA^{-2}\frac {\nabla \rho}{\rho}(A^{-1}x) = W_V(W_Q^\top W_K)^{-1}\frac {\nabla \rho}{\rho}(A^{-1}x). 
$$
In other words, 
$$
\overline{T}^{\infty}_{\mu, \e} \to W_V(W_Q^\top W_K)^{-1}\frac {\nabla \rho}{\rho},  
$$
which is exactly what we wanted to show. Note that when $W_V = - W_Q^\top W_K$ this gives the expected result. 
The general form for the PDE is then 
$$
\partial_t \rho = \mathrm{div}(-W_V(W_Q^\top W_K)^{-1}\frac {\nabla \rho}{\rho} \times \rho)
$$
which gives
$$
\partial_t \rho = \Delta \rho
$$
if $W_V = - W_Q^\top W_K$.
\end{proof}
\subsection{Transformer's PDE - Proof of Proposition \ref{prop:soft}}
\begin{proof}
Let $x \in \RR^d$ and consider 
$$
g_{\e}(x) = \e T^{1}_{\mu,\e}(W^{-1}_Qx) = \frac{\int e^{-\frac{\|x - W_Kx' \|^{2}}{2 \e}}W_Vx'\rho(x')dx'}{\int e^{-\frac{\|x - W_Kx' \|^{2}}{2 \e}}\rho(x')dx'}.
$$
We perform the change of variable $y = W_Kx'$. This gives: 
$$
g_{\e}(x) = \frac{\int e^{-\frac{\|x - y \|^{2}}{2 \e}}W_VW^{-1}_Ky\rho(W^{-1}_Ky)dy}{\int e^{-\frac{\|x - y' \|^{2}}{2 \e}}\rho(W^{-1}_Ky)dy}.
$$
Using the Laplace expansion result from \cite{singer2006graph}, we obtain that
$$
g_{\e}(x) = W_VW^{-1}_K \frac{x \rho(W^{-1}_Kx) + \frac{\e}{2}\Delta(x\rho(W^{-1}_Kx)) + o(\e)}{\rho(W^{-1}_Kx) + \frac{\e}{2}\Delta(\rho(W^{-1}_Kx)) + o(\e) .}
$$
By doing a Taylor expansion for the denominator, we find 
$$
g_{\e}(x) = W_VW^{-1}_K (x +  \frac{\e}{2}\frac{\Delta(x\rho(W^{-1}_Kx)}{\rho(W^{-1}_Kx)} +  o(\e))(1- \frac{\e}{2}\frac{\Delta(\rho(W^{-1}_Kx))}{\rho(W^{-1}_Kx)} + o(\e))
$$
and 
$$
g_{\e}(x) = W_VW^{-1}_K (x +  \e \frac{\nabla(\rho(W^{-1}_Kx))}{\rho(W^{-1}_Kx)} + o(\e).)
$$
Since $\overline{T}^{1}_{\mu, \e} = {T}^{1}_{\mu, \e} + \frac{1}{\e} W_Q^{\top} W_Q = \frac{1}{\e}(g_{\e}(W_Qx) + W_Q^{\top} W_Qx)$ and because $W_VW^{-1}_K = -{W_Q}^{\top}W_KW^{-1}_K = -W^{\top}_Q$ we have 
$$
\overline{T}^{1}_{\mu, \e} = -W^{\top}_Q W^{-1}_K \frac{\nabla \rho(W^{-1}_K W_Qx))}{\rho(W^{-1}_K W_Qx)} + o(1)
$$
which is exactly the expected result. 
\end{proof}
\section{Implementation details}\label{app:imp_details}
We implement Sinkhorn's algorithm in log domain for stability. 
Given a matrix $K^{0} \in \RR^{n \times n}$ such that $K^{0}_{i,j} = e^{C_{i,j}}$ for some $C \in \RR^{n \times n}$, Sinkhorn's algorithm \eqref{eq:sinkhorn_alg} approaches $(f_{},g_{})\in \RR^{n} \times \RR^{n}$ such that $K^{\infty} = \mathrm{diag}(e^{f^{\infty}}) K^0 \mathrm{diag}(e^{g^{\infty}})$ by iterating in $\log$ domain, starting from $g_{}^{0} = \mathbb{0}_n$, 
\begin{equation}
\begin{split}
    f_{}^{l+1} &=  \log(\mathbb{1}_n/n) -  \log({K e^{g_{}^{l}}}) \quad\text{if $l$ is even} \\
    g_{}^{l+1} &=  \log(\mathbb{1}_n/n) -  \log({K^{\top} e^{f_{}^{l}}}) \quad\text{if $l$ is odd}.
\end{split}
\end{equation}

This allows for fast and accurate computations, where $\log({K e^{g_{}^{l}}})$ and $\log({K^{\top} e^{f_{}^{l}}})$ are computed using $\texttt{log-sum-exp}$.

\section{Experimental details}\label{app:exp_details}

\subsection{ModelNet 40 classification}

\paragraph{Set Transformers.}
For our experiments on ModelNet using Set Transformers, we first prepossess the ModelNet 40 dataset. We then uniformly sample $5000$ points from each element in the dataset. Our architecture is composed of two ISAB layers in the encoder and a decoder composed of a SAB and a Pooling by Multihead
Attention (PMA) module.
For the training, we use a batch-size of $64$ and we use Adam \citep{kingma2014adam}. The training is done over $300$ epochs. The initial learning rate is $10^{-3}$ and is decayed by a factor $10$ after $200$ epochs. 

\paragraph{Point Cloud Transformers.}
For our experiments on ModelNet using Point Clouds Transformers, we uniformly sample $1024$ points from each element in the dataset. For the training, we use a batch-size of $32$ and we use SGD \citep{ruder2016overview}. The training is done over $300$ epochs. The initial learning rate is $10^{-4}$ and is decayed by a factor $10$ after $250$ epochs. 

\subsection{Sentiment Analysis}
We use the code available at the repository nlp-turorial\footnote{\href{url}{https://github.com/lyeoni/nlp-tutorial/tree/master/text-classification-transformer}}, where a pretrained Transformer is fine-tuned on the IMDb dataset. In our experiment, we reset the parameters of the pretrained Transformer and train it from scratch on the IMDb dataset. We use an architecture of depth $6$, with $8$ heads. For the training, we use a batch-size of $32$ and we use 
Adam. The training is done over $15$ epochs. The initial learning rate is $10^{-4}$ and is decayed by a factor $10$ after $12$ epochs. 

\subsection{Neural Machine Translation}
We use the Transformer from \texttt{fairseq} and the command for training it on the IWSLT'14\footnote{\href{url}{https://github.com/pytorch/fairseq/blob/main/examples/translation/README.md}} dataset. When fine-tuning a Sinkformer, we simply divide the original learning rate by $10$.

\subsection{Vision Transformers}

\paragraph{Cats and dogs classification.}

This experiment is done on the cats and dogs\footnote{\href{url}{https://www.kaggle.com/c/dogs-vs-cats/data}} dataset. For this experiment, we use a batch-size of $64$ and Adam. We use an architecture of depth $6$, with $8$ heads, and select a patch-size of $16$. The training is done over $300$ epochs. The initial learning rate is $5 \times 10^{-5}$ and divided by $10$ after $250$ epochs. 

\paragraph{Impact of the patch size on the final accuracy.} 

For this experiment, we use a batch-size of $100$ and Adam. We use an architecture of depth $1$, with $1$ heads, without non-linearity, and select different values for the patch-size. The training is done over $45$ epochs. The initial learning rate is $1 \times 10^{-3}$ (resp. $2 \times 10^{-3}$) for the Transformer (resp. Sinkformer) and divided by $10$ after $35$ epochs and again by $10$ after $41$ epochs. 

\end{document}